\documentclass{article}
\usepackage{spconf,amsmath,epsfig}
\usepackage{graphicx}
\usepackage{epstopdf}
\usepackage{booktabs}
\usepackage{caption}
\usepackage{float}
\usepackage{amsmath}
\usepackage{algorithm}
\usepackage{algorithmic}

\newtheorem{myDef}{Definition}
\DeclareMathOperator*{\argmin}{\argmin}

\newcommand{\MYWHILE}{\item[\mywhile]}
\newcommand{\mywhile}{\textbf{Repeat}}

\begin{document}\sloppy

\def\x{{\mathbf x}}
\def\L{{\cal L}}

\title{Deep Boosting: Layered Feature Mining for General Image Classification}
%
\name{Zhanglin Peng$^1$, ~ Liang Lin$^{1,2,*}$\thanks{$^*$Corresponding author is Liang Lin. This work was supported by the Hi-Tech Research and Development Program of China (no. 2013AA013801), Guangdong Science and Technology Program (no. 2012B031500006), Special Project on Integration of Industry, Education and Research of Guangdong Province (no. 2012B091000101), and the open funding project of State Key Laboratory of Virtual Reality Technology and Systems, Beihang University (Grant No. BUAA-VR-12KF-06).}, ~Ruimao Zhang$^1$, ~ Jing Xu$^1$}
\address{$^1$Sun Yat-Sen University, Guangzhou, China\\ $^2$SYSU-CMU Shunde International Joint
Research Institute, Shunde, China.\\ \small{z.l.peng1990@gmail.com, linliang@ieee.org, r.m.zhang1989@gmail.com, xjintw@hotmail.com} }

\maketitle

\begin{abstract}
Constructing effective representations is a critical but challenging problem in multimedia understanding. The traditional handcraft features often rely on domain knowledge, limiting the performances of exiting methods. This paper discusses a novel computational architecture for general image feature mining, which assembles the primitive filters (i.e. Gabor wavelets) into compositional features in a layer-wise manner. In each layer, we produce a number of base classifiers (i.e. regression stumps) associated with the generated features, and discover informative compositions by using the boosting algorithm. The output compositional features of each layer are treated as the base components to build up the next layer. Our framework is able to generate expressive image representations while inducing very discriminate functions for image classification. The experiments are conducted on several public datasets, and we demonstrate superior performances over state-of-the-art approaches.
\end{abstract}
\section{Introduction}
\label{sec:intro}

\begin{figure}[!ht]
\centering
\includegraphics[width=3.5in]{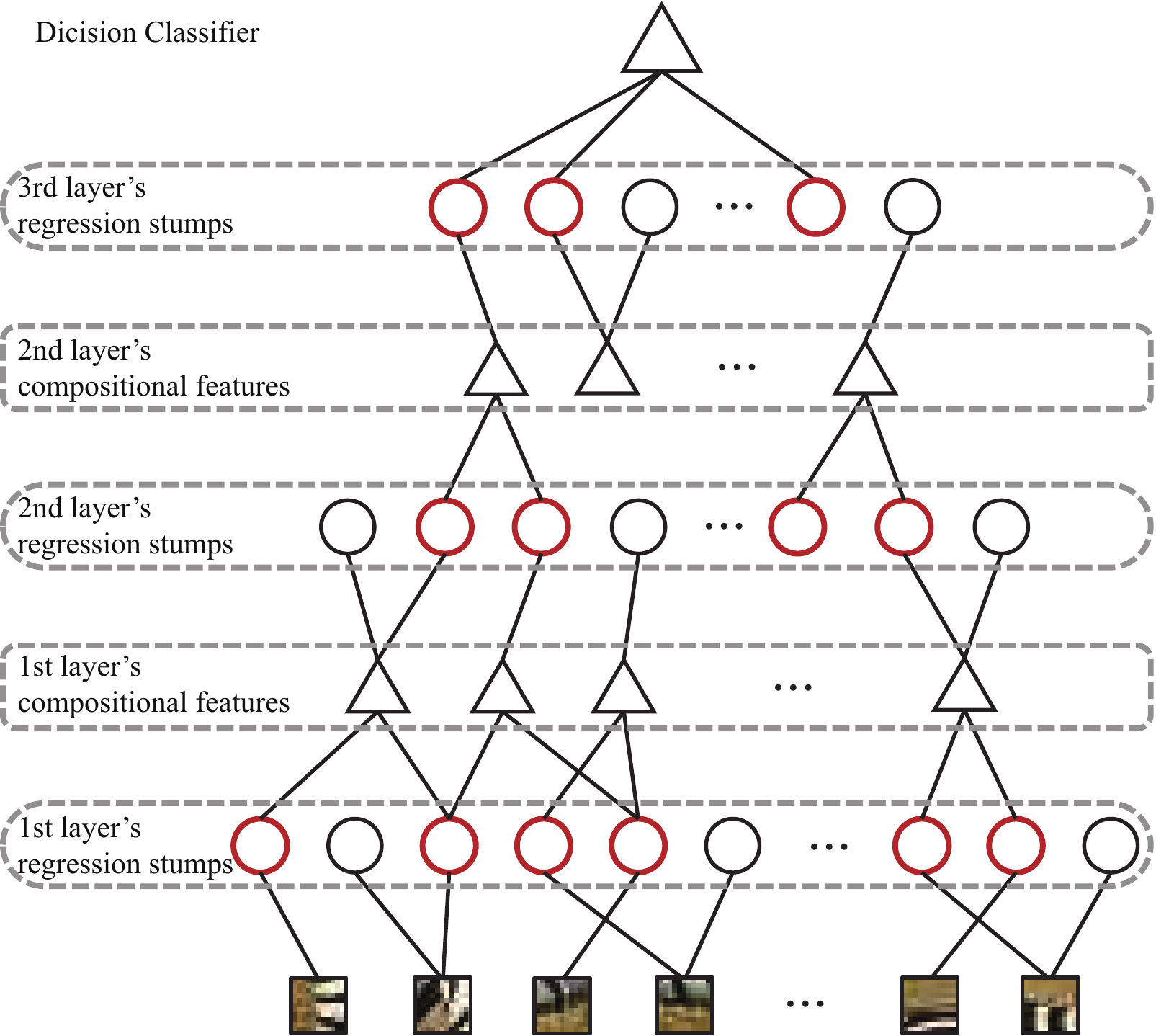}
\caption{Illustration of layered feature mining for deep boosting. Each patch on the bottom denotes an absolute position in the image. Each layer of the deep boosting model except the last one comprises two stages: feature selection and composition. In feature selection stage, the black circles indicate the visual primitive candidates in each layer, the selected features are marked as red. In composition stage, the compositional features are indicated by triangles which is the weighted linear combination of two selected features in the lower layer. At the highest layer, we employ all the final composition features to train the strong classifier to predict the class label of the query image.}
\label{fig:framework}
\end{figure}

 Feature engineering (i.e. constructing effective image representation) has been actively studied in machine learning and computer vision  \cite{FeatureSelection1,FeatureMining,lin2012representing} . In literature, the terms feature selection or feature mining often refer to selecting a subset of relevant feature from a special feature space  \cite{FeatureSelection1,FeatureSelection2,FeatureMining}. One of the typical feature selection method is Adaboost algorithm, which merge the feature selection together with the learning procedure. According to previous work in \cite{FaceDetection}, Adaboost constructs a pool of features (\textit{i.e.} weak classifier) and selects the discriminative ones to form the final strong classifier. These boosting-based approaches provide an effective way for image classification task and achieve outstanding results in the past decade.

Despite the admitted success, such boosting methods are suffered from two essential problems. First, the weak classifier selected at each boosting step is limited by their own discriminative ability when faces with complex classification problems. In order to decrease the training error, the final classifier is linearly combined by a large numbers of weak classifiers through boosting \cite{MiningCompositionalFeatures}. On the other hand, amounts of effective learning procedure always lead the training error approaching to zero. However, under the unknown decision boundary, how to decrease the test error when training error is approaching zero is still an open issue \cite{Gentle-Adaboost}.

In recent decades, the hierarchical models, also known as deep models \cite{CNN,DBN} have played an irreplaceable role in multimedia and computer vision literature. Generally, such hierarchical architecture represents different layer of vision primitives such as pixels, edges, object parts and so on. The basic principles of hierarchical models are concentrated on two folds:  (1) layerwise learning philosophy, whose goal is to learn single layer of the model individually and stack them to form the final architecture; (2) feature combination rules, which aim at utilizing the combination of low layer detected features to construct the high layer impressive features by introducing the activation function. In this paper, the related exciting researches inspire us to employ such compositional representation to construct the impressive features with more discriminative power. Different from previous works \cite{CNN,DBN,CDBN} applying the hierarchical generative model, we address the problem on general image classification directly and design the final classifier leveraging the generalization and discrimination abilities.

This paper proposes a novel feature mining framework, namely \textit{deep boosting}, which aims to construct the effective discriminative features for image classification task. Compared with the concept 'mining' proposed in \cite{FeatureMining}, whose goal is picking a subset of features as well as modeling the entire feature space, we utilize the word to describe the processing of feature selection and combination, which is more related to \cite{MiningCompositionalFeatures}. For each layer, following the famous boosting method \cite{Gentle-Adaboost}, our deep model sequentially selects visual features to learn the classifier to reduce the training error. In order to construct high-level discriminative representations, we composite selected features in the same layer and feed into higher layer to build a multilayer architecture. Another key to our approach is introducing the spatial information when combining the individual features, that inspires upper layer representation more structured on the local scale. The experiment shows that our method achieves excellent performance on image classification task.


\section{Related Work}

In the past few decades, many works focus on designing different types of features to capture the characteristics of images such as color, SIFT and HoG~\cite{HOG}.  Based on these feature descriptors, Bag-of-Feature (BoF) model seems to be the most classical image representation method in computer vision and related multimedia applications. Several promising studies \cite{15Scenes,ScSPM,LLC} were published to improve this traditional approach in different aspects. Among these extension, a class of sparse coding based methods \cite{ScSPM,LLC}, which employ spatial pyramid matching kernel (SPM) proposed by Lazebnik \textit{et al}, has achieved great success in image classification problem. Despite we are developing more and more effective representation methods, the lack of high-level image expression still plagues us to build up the ideal vision system.

On the other hand, learning hierarchical models to simultaneously construct multiple levels of visual representation has received much attention recently~\cite{lin2009stochastic}. Our deep boosting method is partially motivated by recent developed deep learning techniques~\cite{CNN,DBN,SumProductNetwork}. Different from previous hand-craft feature design method, deep model learns the feature representation from raw data and validly generates the high-level semantic representation. However, as shown in recent study \cite{SumProductNetwork}, these network-based hierarchical models always contain thousands of nodes in a single layer, and is too complex to control in real multimedia application. In contrast, an obvious characteristic of our study is that we build up the deep architecture to generate expressive image representation simply and obtains the near optimal classification rate in each layer.


\section{Deep Boosting for Image Recognition}


\subsection{Background: Gentle Adaboost}

We start with a brief review of Gentle Adaboost algorithm \cite{Gentle-Adaboost}. Without loss of generality, considering the two-class classification problem, let $(x_1,y_1)...(x_N,y_N)$ be the training samples, where $x_i$ is a feature representation of the sample and $y_i\in \{-1,1\}$. $w_i$ is the sample weight related to $x_i$. Gentle Adaboost \cite{Gentle-Adaboost,SharingFeatures} provides a simple additive model with the form,
\begin{equation}\label{eq1}
F(x_i) = \sum_{m=1}^{M} f_m(x_i),
\end{equation}
where $f_m$ is called weak classifier in the machine learning literature. It often defines $f_m$  as the regression stump $f_m(x_i)= a\hbar(x_i ^{d}>\delta)+b$, where $\hbar(\cdot)$ denotes the indicator function, $x_i ^{d}$ is the $d$-th dimension of the feature vector $x_i$, $\delta$ is a threshold, $a$ and $b$ are two parameters contributing to the linear regression function. In iteration $m$, the algorithm learns the parameter $(d,\delta,a,b)$ of $f_m(\cdot)$ by weighted least-squares of $y_i$ to $x_i$ with weight $w_i$,
\begin{equation}\label{eq2}
\min_{1\leq d \leq D} \sum_{i=1}^N w_i\parallel a^{d}\hbar(x_i^{d} >\delta^{d})+b^{d} - y_i  \parallel ^2 ,
\end{equation}
where $D$ is the dimension of the feature space. In order to give much attention to the cases that are misclassified in each round, Gentle Adaboost adjusts the sample weight in the next iteration as $w_i\leftarrow w_i e^{-y_i f_m(x_i)}$ and updates $F(x_i) \leftarrow F(x_i)+f_m(x_i)$. At last, the algorithm outputs the result of strong classifier as the form of sign function $sign[F(x_i)]$. Please refer to \cite{Gentle-Adaboost,SharingFeatures} for more academic details.

\begin{figure}[!ht]
\centering
\includegraphics[width=3.3in]{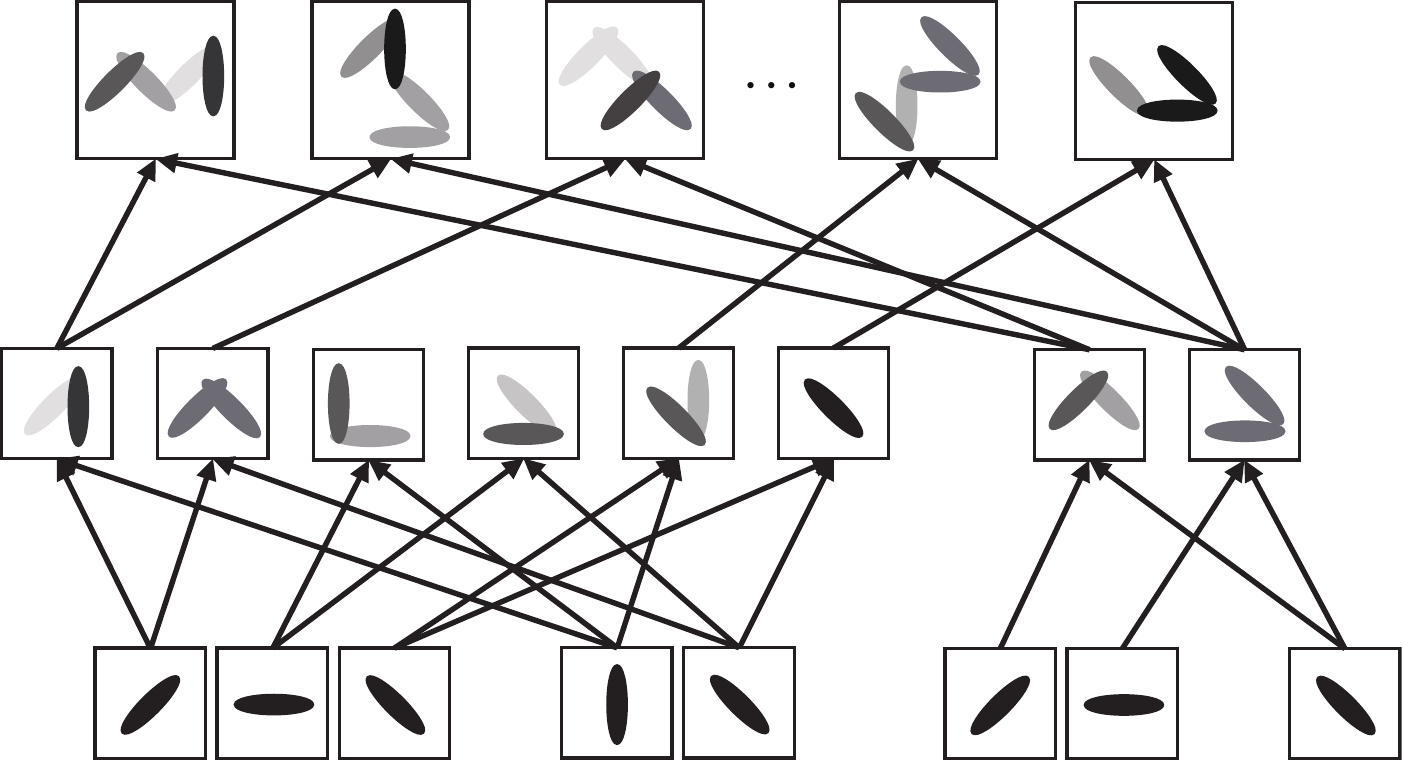}
\caption{Illustration of Feature Combination. A cluster on the bottom denotes a set of different selected visual primitives (\textit{i.e.} Gabor wavelet filters) at the same position in the image. A Gabor wavelet filter is denoted by an ellipse. At the second layer, a composite feature, which is combined by two Gabor wavelet filters, is fed into the third layer as an upper visual primitive. The intensity of every ellipse indicates the weight of Gabor wavelet filter.}
\label{fig:FeatureCombine}
\end{figure}

\subsection{Preprocessing}
\label{sec:Preprocessing}

The basic units in the Gentle Adaboost algorithm are individual features, also known as weak classifiers. Unlike the rectangle feature in \cite{FaceDetection} for face detection, we employ Gabor wavelets response as the image feature representation. Let $I$ be an image defined on image lattice domain and $G$ be the Gabor wavelet elements with parameters $(w,h,\alpha,s)$, where $(w,h)$ is the central position belonging to the lattice domain, $\alpha$ and $s$ denote the orientation and scale parameters. Following \cite{ActiveBasisModel}, we utilize the normalized term to make the Gabor responses comparable between different training images:
\begin{equation}\label{eq3}
\xi^2(s) = \frac{1}{|P|A}\sum_{\alpha}\sum_{w,h} |\langle I,G_{w,h,\alpha,s}\rangle|^2 ,
\end{equation}
where $|P|$ is the total number of pixels in image $I$, and $A$ is the number of orientations. $\langle\cdot\rangle$ denotes the convolution process. For each image $I$, we normalize the local energy as $|\langle I,G_{w,h,\alpha,s}\rangle|^2 / \xi^2(s)$ and define positive square root of such normalized result as feature response. In practice, we resize image into $120\times120$ pixels and apply one scale and eight orientations in our implementation, so there are total $120\times120\times1\times8$ filter responses for each grayscale image.

\subsection{Discriminative Feature Selection}
\label{sec:selection}

In this subsection, we set up the relationship between the weak classifier and Gabor wavelet representation. After the Gabor responses calculated, we learn the classification function utilizing the given feature set and the training set including both positive and negative images. Suppose the size of the training set is $N$. In our deep boosting system, the weak learning method is to select the single feature ( \textit{i.e.} weak classifier ) which best divides the positive and negative samples. To fix the notation, let $x_i \in R^D$ be the feature representation of image $I_i$, where $D$ is the dimension of the feature space. It is obvious that $D = 120\times120\times1\times8$ in the first layer, corresponding to Gabor wavelets in Sec.(\ref{sec:Preprocessing}). Specifically, each element of $x_i$ is a special Gabor response of image $I_i$ (in the first layer) or their composition (in other layers). Note that in the rest of the paper, we apply $x_i^d$ to denote the value of $x_i$ in the $d$-th dimension. In each round of feature selection procedure, instead of using the indictor function in Eq.(\ref{eq2}), we introduce the sigmoid function defined by the formula:

\begin{equation}\label{eq-ex1}
\phi(x) = 1/(1+e^{-x})
\end{equation}
 In this way, we consider a collection of regressive function $\{f^1,f^2,...,f^D\}$ where each $f^d$ is a candidate weak classifier whose definition is given in Definition. \ref{def:FeatureSelection}.

\begin{myDef}[Discriminative Feature Selection]
\label{def:FeatureSelection}
\textrm{\\}
In each round, the algorithm retrieves all of the candidate regression functions, each of which is formulated as:
\begin{equation}\label{eq4}
f^d(x_i)= a\phi(x_i^{d} - \delta)+b ,
\end{equation}
where $\phi(\cdot) $ is a sigmoid function defined in Eq.(\ref{eq-ex1}). The candidate function with current minimum training error is selected as the current weak classifier $f$, such that
\begin{equation}\label{eq5}
\min_{d} \sum_{i=1}^N w_i\parallel f^d(x_i) - y_i  \parallel ^2 ,
\end{equation}
where $f^d(x_i)$ is associate with the $d$-th element of $x_i$ and the function parameter $(\delta,a,b)$.
\end{myDef}

According to the above discussion, we build the bridge between the weak classifier and the special Gabor wavelet ( or their composition ), thus the weak classifiers learning can be viewed as the feature selection procedure in our deep boosting model.


\subsection{Composite Feature Construction}
\label{sec:composite}

Since the classification accuracy based on an individual feature or single weak classifier is usually low and the strong classifier, which is the weighted linear combination of weak classifiers, is hardly to decease the test error when training error is approaching to zero. It is of our interest to improve the discriminative ability of features and learn high-level representations as well.

In order to achieve the goal above, we introduce the feature combination strategy in Definition.\ref{def:FeatureComposite}. All features selected in the feature selection stage are combined in a pair-wise manner with spatial constraints, and the output composition features of each layer are treated as base components to construct the next layer.

\begin{myDef}[Feature Combination Rule]
\label{def:FeatureComposite}
\textrm{\\}
For each image $I$, whose feature representation is denoted by $x$, we combine two selected features in local area as,
\begin{equation}\label{eq6}
[x^j]_{l+1} = \beta_s \: [x^s]_{l} + \beta_t \: [x^t]_{l}  \:\:\:\:\:\:  \exists s,t \in \Omega(j)
\end{equation}
where  $[x^s]_{l}$ and $[x^t]_{l}$ indicate the $s$-th and $t$-th feature response corresponding to the image $I$ in the layer $l$.
\end{myDef}

As illustrate in the Fig.(\ref{fig:framework}), $x^s$ and $x^t$ are response values of selected features which are indicated by the red circles in each layer. $\beta_s$ and $\beta_t$ are the combination weights proportion to the training error rates of $s$-th and $t$-th weak classifiers calculated over the training set. $\Omega(j)$ is the local area determined by the projection coordinate of composition feature $j$ on the normalized image ( \textit{i.e.} the image with the size of $120\times 120$ pixels in practice ). In the higher layer, the feature selection process is the same as the lower layer, which can be formulated as Eq.(\ref{eq5}). Please refer to Fig.(\ref{def:FeatureComposite}) for more details about feature combination.

Integrating the two stages in Sec.(\ref{sec:selection}) and Sec.(\ref{sec:composite}), we build up the single layer of our model. Then we stack them to form the final deep boosting architecture which consist of many layers. The overall of our feature mining algorithm is summarized in Algorithm(\ref{alg:Framwork}).

\begin{algorithm}[!ht]
\caption{Deep Boosting for Feature Mining}
\label{alg:Framwork}
\begin{algorithmic}
\REQUIRE ~~\\
    Positive and negative training samples $(x_1,y_1)...(x_N,y_N)$, the number of selected features $M_l$ in layer $l$, the total layer number $L$.
\ENSURE ~~\\                           
    A pool of generated features $\Psi$ and the final classifier $F^L(x)$ for a special category.
\MYWHILE for $l=1,2,\ldots,L$:
    \STATE
    \begin{itemize}
    \setlength{\itemsep}{1pt}
    \setlength{\parskip}{3pt}
    \setlength{\parsep}{10pt}
    \item[1.] Start with score $F^l(x)=0$ for layer $l$ and sample weights $w_i=1/N$, $i=1,2,\ldots,N$.
    \item[2.] Select features and learn the strong classifier for layer $l$ as follows:

        \textbf{Repeat} for $m=1,2,\ldots,M^l$:
        \setlength{\parskip}{-1pt}
            \begin{itemize}
            \setlength{\itemsep}{1pt}
            \setlength{\parskip}{0pt}
            \setlength{\parsep}{10pt}
            \item[(a)] Learn the current weak classifier $f_m$ by Eq.(\ref{eq5}).
            \item[(b)] Update $w_i\leftarrow w_i e^{-y_i f_m(x)}$ and renormalize.
            \item[(c)] Update $F^l(x) \leftarrow F^l(x)+f_m(x)$.
            \end{itemize}
        \setlength{\parskip}{0pt}

    \item[3.] Update $\Psi$ by $f_m(x)$, $m=1,2,\ldots,M^l$.
    \item[4.] Generate the composite features according to Eq.(\ref{eq6}).
    \end{itemize}
\end{algorithmic}
\end{algorithm}

\subsection{Multi-class Decision}

We employ the naive \textit{one-against-all} strategy to handle the multi-class classification task in this paper. Given the training data $\{(x_i,y_i)\}_{i=1}^N$,$y_i\in \{1,2,...,K\}$, we train $K$ binary strong classifiers, each of which returns a classification score for a special test image. In the testing phrase, we predict the label of image referring to the classifier with the maximum score.

\begin{table*}[tb]
\footnotesize
\centering
\caption{Classification Rate(\%) on Caltech256 Class Sets - Easy10.}
\begin{tabular}{lccccccccccc}
\toprule
            &desk-globe&mars&sheet-music&sunflower& tower-pisa & trilobite & watch & zebra & car-side & face-easy & \textbf{AVERAGE}\\
\midrule
ScSPM \cite{ScSPM}                  &92.31&88.10&87.04&96.10&\textbf{100.0}&81.25&90.06&\textbf{93.90}&\textbf{100.0}&\textbf{100.0}&92.87\\
  LLC \cite{LLC}                    &92.30&88.09&81.48&\textbf{100.0}&96.66&93.75&\textbf{92.98}&90.90&\textbf{100.0}&\textbf{100.0}&93.61\\
HoG+SVM \cite{HOG}                  &89.09&81.45&64.16&85.50&71.66&80.29&84.75&77.20&99.28&98.24&83.16\\
\textbf{Ours}                       &\textbf{100.0}&\textbf{93.75}&\textbf{91.66}&\textbf{100.0}&96.66&\textbf{97.05}&82.97&88.90&\textbf{100.0}&98.93&\textbf{94.99}\\
\bottomrule
\label{Tab:Easy10}
\end{tabular}
\end{table*}

\begin{table*}[tb]
\footnotesize
\centering
\caption{Classification Rate(\%) on  Caltech256 Class Sets - Var10.}
\begin{tabular}{lccccccccccc}
\toprule
            & bear & billiards & blimp & hamburger & hummingbird & laptop & minotaur & roulette & skyscraper & yo-yo & \textbf{AVERAGE}\\
\midrule
ScSPM \cite{ScSPM}       & 80.55 & \textbf{78.22} & 64.28 & 76.78 & 62.79 & 63.26 & 59.61 & 62.26 & 87.69 & 65.71 & 70.11\\
LLC \cite{LLC}         & 79.16 & 74.19 & 69.64 & 78.57 & 67.44 & 70.40 & 73.07 & 56.60 & 86.15 & 64.28 & 71.95\\
HoG+SVM \cite{HOG}     & \textbf{88.80} & 74.49 & 74.23 & 81.15 & \textbf{84.46} & \textbf{83.23} & 79.09 & 69.13 & 79.99 & 65.24 & 77.98\\
\textbf{Ours}& 88.09 & 62.38 & \textbf{92.30} & \textbf{96.15} & 83.92 & 80.88 & \textbf{81.81} & \textbf{100.0} & \textbf{91.42} & \textbf{77.50} & \textbf{85.45}\\
\bottomrule
\label{Tab:Var10}
\end{tabular}
\end{table*}

\begin{figure}[!ht]
\centering
\includegraphics[width=3.4in]{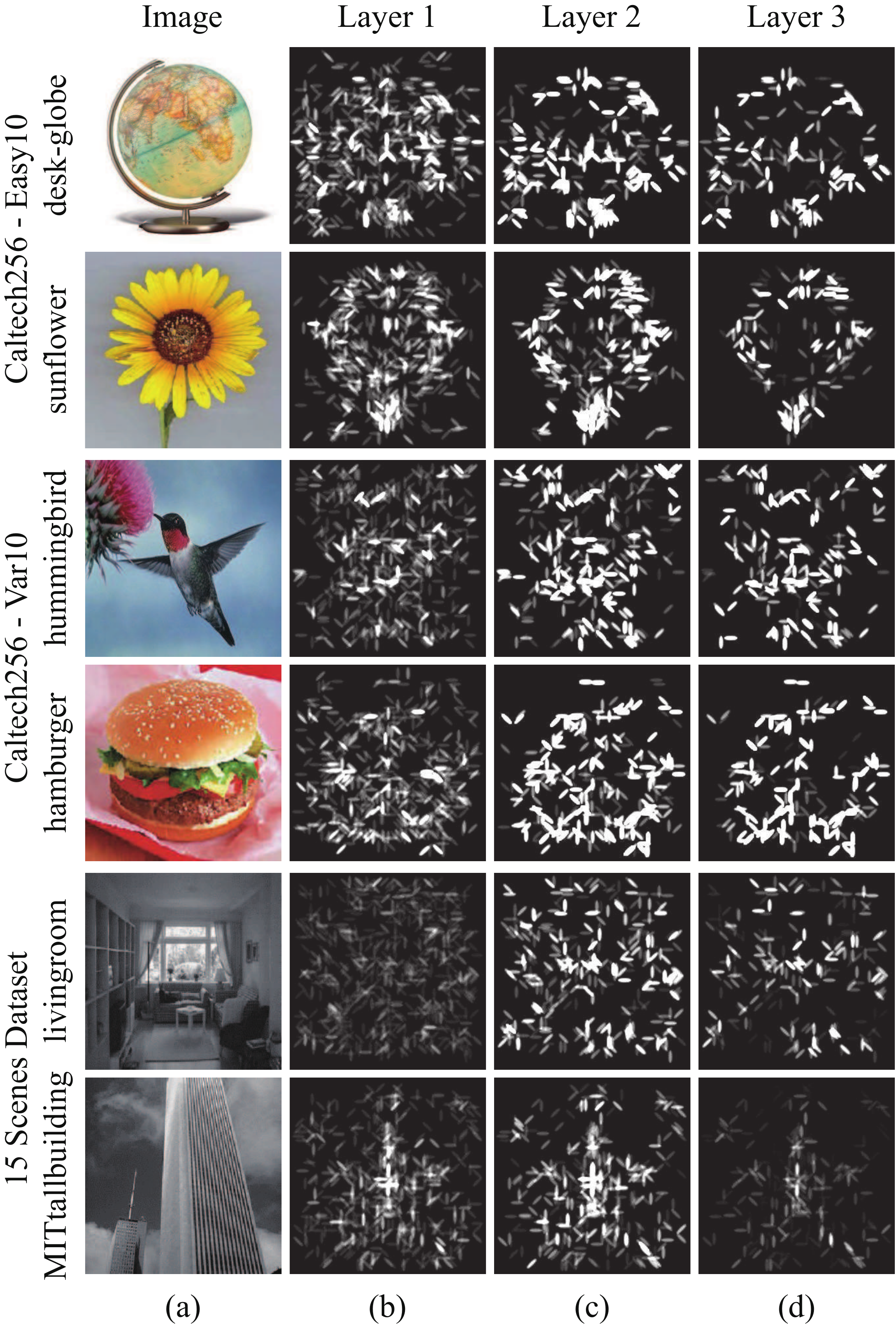}
\caption{Visualizations of Deep Boosting. (a) Original image; (b) Visualizations of the $1^{st}$ layer; (c) Visualizations of the $2^{nd}$ layer; (d) Visualizations of the $3^{rd}$ layer. Elliptical bars in each figure denote Gabor wavelets, and the shade of color shows the corresponding weight. }
\label{fig:Results}
\end{figure}

\section{Experiment}

\subsection{Dataset and Experiment Setting }

We apply the proposed method on general classification task, using Caltech 256 Dataset \cite{Caltech256} and the 15 Scenes Dataset \cite{15Scenes} for validation. For both datasets, we split the data into training and test, utilize the training set to discover the discriminative features and learn the strong classifiers, and apply the test to evaluate classification performance.

As mentioned in Sec.(\ref{sec:Preprocessing}). For both datasets, we resize each image as $120\times120$ pixels, and simply set the Gabor wavelets with one scale and eight orientations. In each layer, the strong classifier training is performed in a supervised manner and the number of selected features are set as 1000, 800, 500 respectively. We combine the selected features in the $3\times3$ block densely and capture $3000\sim8000$ composite features every layer. According to the experiment, the number of composite features in each layer relies on the complexity of image content seriously. The visualization of feature map in each layer is shown in Fig.(\ref{fig:Results}).

We carry out the experiments on a PC with Core i7-3960X 3.30 GHZ CPU and 24GB memory. On average, it takes $5\sim9$ hours for training a special category model, depending on the numbers of training examples and the complexity of image content. The time cost for recognizing a image is around $25\sim40$ seconds.

\subsection{Experiment I: Caltech 256 Dataset }

We evaluate the performance of our deep boosting algorithm on the Caltech 256 Dataset \cite{Caltech256} which is widely used as the benchmark for testing the general image classification task \cite{ScSPM,LLC}. The Caltech 256 Dataset contains 30607 images in 256 categories. We consider the image classification problem on Easy10 and Var10 image sets according to \cite{LargeScaleOnlineLearning}. We evaluate classification results from 10 random splits of the training and testing data ( \textit{i.e.} 60 training images and the rest as testing images ) and report the performance using the mean of each class classification rate. Besides our own implementations, we refer some released Matlab code from previous published literature \cite{ScSPM,LLC} in our experiments as well. As Tab.(\ref{Tab:Easy10}) and Tab.(\ref{Tab:Var10}) report, our method reaches the classification rate of $94.9\%$ and $85.4\%$ on Easy10 and Var10 datasets, outperforming other approaches \cite{HOG,LLC,ScSPM}.

\subsection{Experiment II: 15 Scenes Dataset }

\begin{figure}[!ht]
\centering
\includegraphics[width=2in]{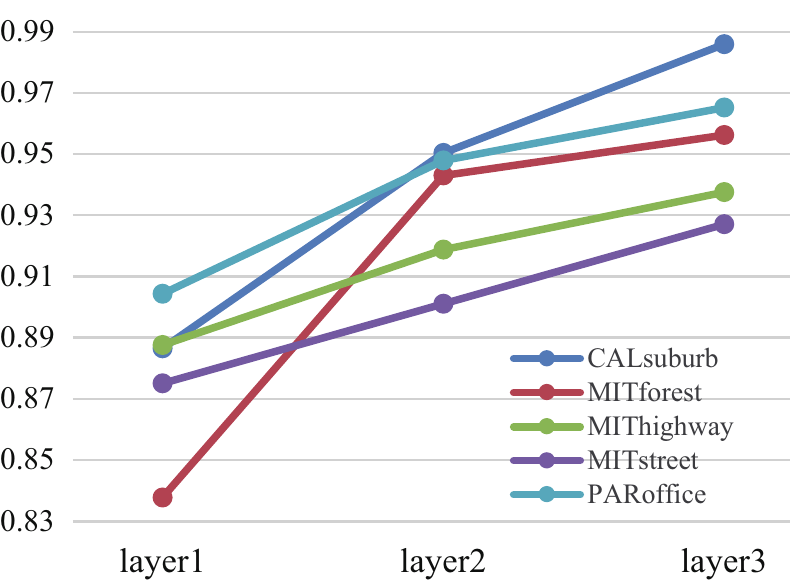}
\caption{Classification accuracy of our proposed deep boosting method applying each layer's strong classifier. We select results from top five categories in 15 Scenes Dataset to report. }
\label{fig:5class}
\end{figure}

We also test our method on the 15 Scenes Dataset \cite{15Scenes}. This dataset totally includes 4485 images collected from 15 representative scene categories. Each category contains at least 200 images. The categories vary from mountain and forest to office and living room. As the standard benchmark procedure in \cite{ScSPM,15Scenes}, we select 100 images per class for training and others for testing. The performance is evaluated by randomly taking the training and testing images 10 times. The mean and standard deviation of the recognition rates are shown in Table(\ref{Tab:15Scenes}). In this experiment, our deep boosting method achieves better performance than previous works \cite{GIST,ScSPM} as well. Note that, instead of HoG+SVM, we compare our approach with GIST+SVM method in this experiment, due to the effectiveness of GIST \cite{GIST} in the scene classification task. Considering the subtle engineering details, we can hardly achieve desired results applying \cite{LLC} and \cite{ScSPM} methods in our own implementations. So we quote the reported result directly from \cite{ScSPM} and abandon \cite{LLC} as a way of comparison. We also compare the recognition rate utilizing different layer's strong classifier, the results of top five outstanding categories on 15 Sences Dataset are reported in Fig.(\ref{fig:5class}). It is obvious that our proposed feature combination strategy improve the performance effectively.

\begin{table}[h]
\centering
\caption{Classification Rate(\%) on 15 Scenes Dataset.}
\begin{tabular}{lc}
\toprule
Algorithm                           & mean Average Precision\\
\midrule
ScSPM \cite{ScSPM}                  &  80.28 $\pm$ 0.93 \\
GIST+SVM \cite{GIST}                &  75.12 $\pm$ 1.27 \\
\textbf{Ours}                       &  \textbf{81.76 $\pm$ 0.97} \\
\bottomrule
\label{Tab:15Scenes}
\end{tabular}
\end{table}

\section{Conclusion}

This paper studies a novel layered feature mining framework named \textit{deep boosting}. According to the famous boosting algorithm, this model sequentially selects the visual feature in each layer and composites selected features in the same layer as the input of upper layer to construct the hierarchical architecture. Our approach achieves the excellent success on several image classification tasks. Moreover, the philosophy of such deep model is very general and can be applied to other multimedia applications.

\bibliographystyle{IEEEbib}
\bibliography{icme2013template}

\end{document}